\documentclass[10pt,journal,compsoc]{IEEEtran}

\usepackage{mdframed}
\newmdenv[topline=false,bottomline=false,rightline=false]{leftonly}

\newcounter{nquestion}

\newcommand{\etal}{\textit{et al}.}

\ifCLASSOPTIONcompsoc
  \usepackage[nocompress]{cite}
\else
  \usepackage{cite}
\fi

\ifCLASSINFOpdf
  \usepackage[pdftex]{graphicx} 
\else
\fi

\usepackage{amsmath}
\usepackage{amssymb} 
\usepackage{siunitx} 
\usepackage{subfigure}
\usepackage{url}
\usepackage[pagebackref=true,breaklinks=true,colorlinks,bookmarks=false]{hyperref}
\usepackage{multirow}
\usepackage{array}
\usepackage{makecell}

\usepackage{color}

\begin{document}

\title{Face-from-Depth for Head Pose Estimation \\ on Depth Images}

\author{Guido~Borghi,
        Matteo~Fabbri,
        Roberto~Vezzani,
        Simone~Calderara
        and~Rita~Cucchiara
\IEEEcompsocitemizethanks{\IEEEcompsocthanksitem The authors are with the Department of Engineering "Enzo Ferrari", University of Modena and Reggio Emilia, Italy.\protect\\
E-mail: name.surname@unimore.it%
}
}


\IEEEtitleabstractindextext{%
\begin{abstract}
Depth cameras allow to set up reliable solutions for people monitoring and behavior understanding, especially when unstable or poor illumination conditions make unusable common RGB sensors.    
Therefore, we propose a complete framework for the estimation of the head and shoulder pose based on depth images only.
A head detection and localization module is also included, in order to develop a complete end-to-end system. 
The core element of the framework is a Convolutional Neural Network, called \textit{POSEidon$^+$}, that receives as input three types of images and provides the 3D angles of the pose as output. 
Moreover, a \textit{Face-from-Depth} component based on a \textit{Deterministic Conditional GAN} model is able to hallucinate a face from the corresponding depth image. We empirically demonstrate that this positively impacts the system performances. 
We test the proposed framework on two public datasets, namely \textit{Biwi Kinect Head Pose} and \textit{ICT-3DHP}, and on \textit{Pandora}, a new challenging dataset mainly inspired by the automotive setup. 
Experimental results show that our method overcomes several recent state-of-art works based on both intensity and depth input data, running in real-time at more than 30 frames per second.
\end{abstract}

}

\maketitle
\IEEEdisplaynontitleabstractindextext
\IEEEpeerreviewmaketitle

\IEEEraisesectionheading{\section{Introduction}\label{sec:introduction}}

\IEEEPARstart{C}{omputer vision} has been addressing the problem of head pose estimation for several years.\\
In 2009, Murphy-Chutorian and Trivedi \cite{Trivedi2009} made a first assessment of the proposed techniques. More recently, different approaches have been proposed together with some annotated datasets useful for both training and testing those systems. The interest of the research community is mainly due to a large number of applications that require or are improved by a reliable head pose estimation: face recognition with aliveness detection, human-computer interaction, people behavior understanding are some examples. 
Moreover, a large effort has been recently devoted to applications in the automotive field, such as monitoring drivers and passengers. Together with the estimation of the upper-body and shoulder pose, the head monitoring is one of the key technologies required to set up (semi)-autonomous driving cars, human-car interaction for entertainment, and driver's attention measurement. 

In the automotive field, vision-based systems are required to cooperate or even replace other traditional sensors, due to the increasing presence of cameras inside new car's cockpits and to the ease of capturing images and videos in a completely non-invasive manner. \\
In the past, encouraging results for driver head pose estimation have been achieved using RGB images \cite{Trivedi2009,tran2011,bergasa2006,doshi2012,czuprynski2014} as well as different camera types, such as infrared \cite{ji2004real}, thermal \cite{trivedi2004occupant}, or depth \cite{meyer2015,malassiotis2005,borghi17cvpr}. Among them, the last ones are very promising, since they allow robustness when facing strong illumination variations. Moreover, standard techniques based on RGB images are not always feasible due to poor or absent illumination conditions during the night or to the continuous illumination changes during the day.


\begin{figure}[t!]
     \centering
     \includegraphics[width=0.9\columnwidth]{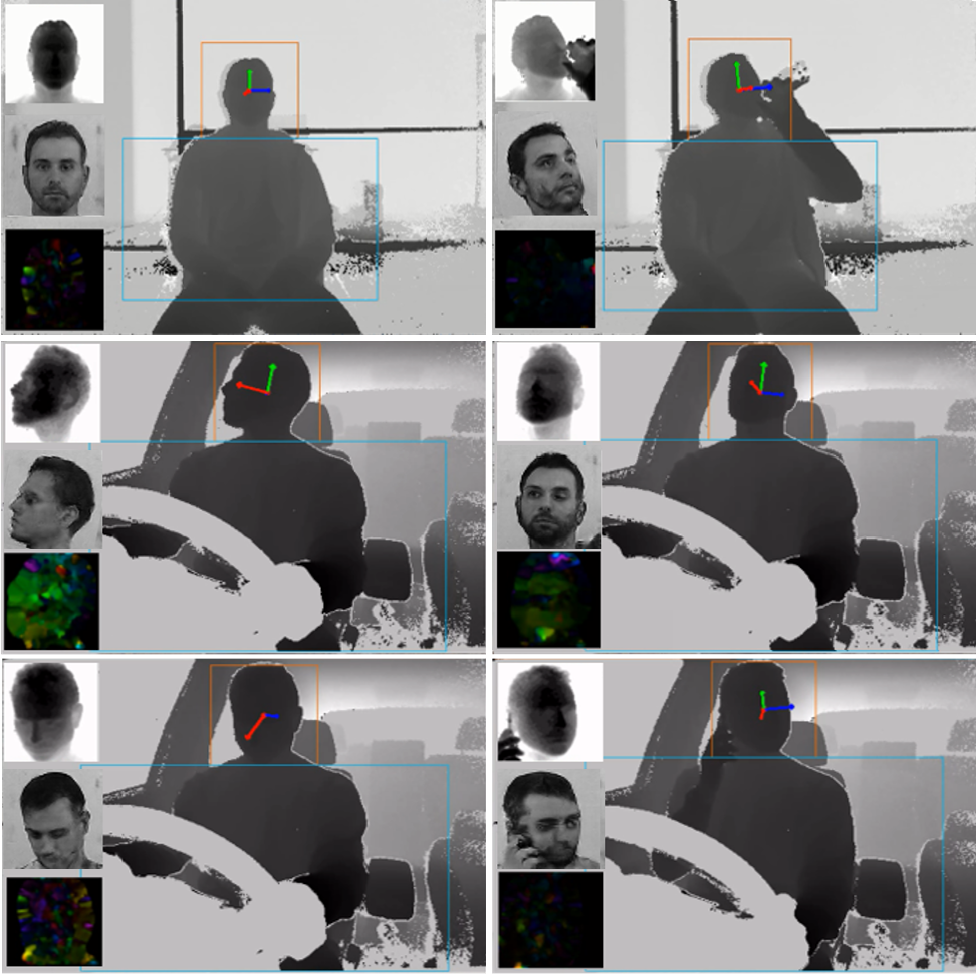}
     \caption{Visual examples of the proposed framework output in indoor (first row) and automotive (second and third row) settings. Head pose angles are reported as colored arrows. Depth maps, \textit{Face-from-Depth} and Motion Image inputs are depicted on the left of each frame.
     }
     \label{fig:demofinale}
 \end{figure}


Nowadays, the acquisition of depth data is feasible thanks to commercial low-cost, high-quality and small-sized depth sensors, that can be easily placed inside the vehicle. 

\begin{figure*}[t!]
    \centering
    \includegraphics[width=0.85\linewidth]{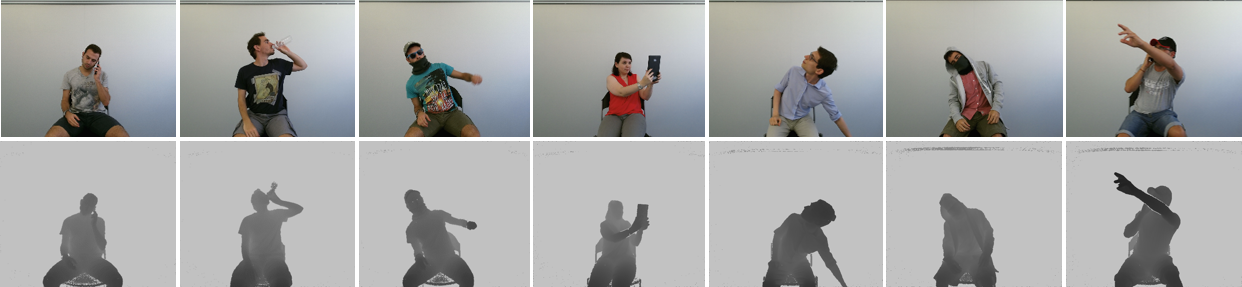}
    \caption{Sample frames from the \textit{Pandora} dataset. As depicted, extreme poses and challenging camouflage can be present.}
    \label{fig:pandora}
\end{figure*}

In this paper, we propose a robust and fast solution for head and shoulder pose estimation, especially devoted to drivers in cars, but that can be easily generalized to any application where depth images are available. 
 \begin{figure}[t!]
     \centering
     \includegraphics[width=0.7\columnwidth]{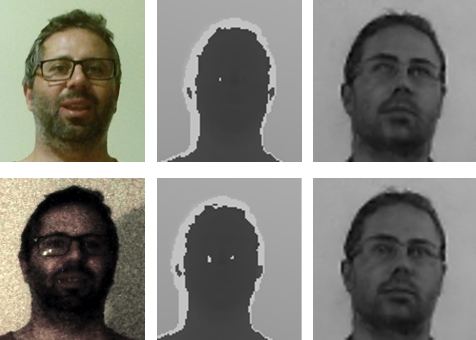}
     \caption{Example of reliability of the \textit{FfD} network on depth images. Two consecutive frames have been selected from a sequence with an abrupt illumination change (from light to dark). In the first column the auto equalized RGB, then the corresponding depth maps and finally the \textit{FfD} reconstruction output.
     }
     \label{fig:darkscene}
 \end{figure}
The presented framework provides impressive results, reaching an accuracy higher than 73\% on the new \textit{Pandora} dataset (see Fig. \ref{fig:pandora}) and a low average error on the \textit{Biwi} dataset, thus overcoming all state-of-art related works.

\begin{figure*}[bt]
    \centering
    \includegraphics[width=0.90\linewidth]{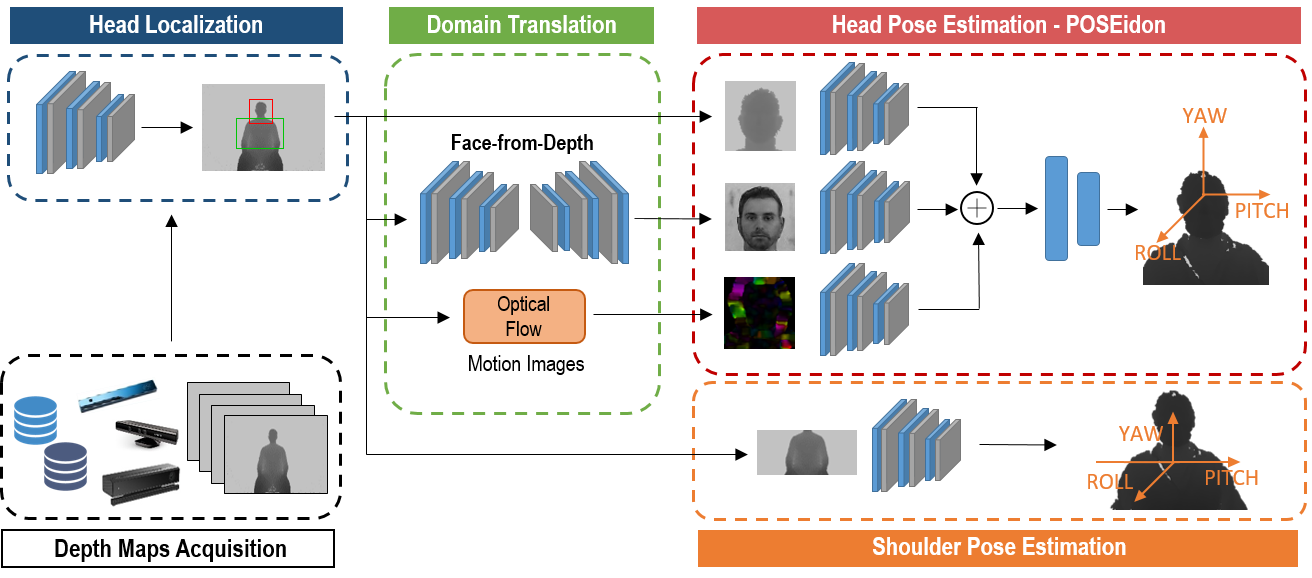}
    \caption{Overview of the whole \textit{POSEidon$^+$} framework. Depth input images are acquired by depth sensors (black) and provided to a head localization CNN (blue) to suitably crop the images around the upper-body or head regions. The head crop is used to produce the three inputs for the following networks (green), that are then merged to output the head pose (red). In particular, the \textit{Face-from-Depth} architecture reconstructs gray-level face images from the corresponding depth maps, while the Motion Images are obtained by applying the \textit{Farneback} algorithm. Finally, the upper-body crop is used for the shoulder pose estimation (orange). [best in color]}
    \label{fig:general} 
\end{figure*}

The core of the framework is a Convolutional Neural Network (CNN), called \textit{POSEidon$^+$}, that combines depth, appearance and Motion Images as input to estimate the 3D pose angles in regression. An overview of the model is depicted in Figure \ref{fig:general}.
The model is enhanced with a \textit{Face-from-Depth (FfD)} component. \\
This is motivated by recent literature results\cite{ahn2014,drouard2015} that testifies the importance of intensity images for the task. The \textit{FfD} component is able to reconstruct the gray-level appearance of a face directly from the corresponding depth image. Thanks to the insensitivity of the depth image to the external illumination conditions, the provided reconstruction is more stable and reliable than gray or color images captured from the same RGB-D sensor. Moreover, the reconstruction can be applied in situations where the depth sensor is exploited alone without the color stream for computational or implementation constraints. \\
As an example, in Figure \ref{fig:darkscene}, we have reported two frames captured from an RGB-D sensor in correspondence of an abrupt illumination change (from light to dark). The depth images are not affected by the illumination change and thus the corresponding \textit{FfD} reconstructions are identical. The provided output highlights the reliability of the developed network as well as the quality of the results.

The overall system is split into two components: the \textit{Face-from-Depth} architecture followed by the pose estimation module, that takes as input the reconstructed gray level images. 
From a first glance, this approach could be improper since we are somehow forcing the \textit{FfD} model to output a human understandable intermediate representation, \textit{i.e.}, the gray level image. Training an end-to-end system enables the network to find the best internal/intermediate representation.  
However, in addition to a performance improvement as reported in Section \ref{sec:results}, the introduction of the \textit{Face-from-Depth} component allows the second part of the system to be trained on wider datasets since more annotated datasets on gray-level images are usually available rather than on depth ones. 
More generally, \textit{FfD} moves input depth images on a domain where more experience is available in order to understand and process them.

This paper is an improved and extended version of our preliminary work, that has been described in \cite{borghi17cvpr}, where the body pose estimation task was carried on through a baseline version of the \textit{POSEidon$^+$} framework, here referred as \textit{POSEidon}. 
In this paper we present the overall framework, introducing a new \textit{Face-from-Depth} architecture, which exploits the recent \textit{Deterministic Conditional GAN} models \cite{isola2016image} to reconstruct gray-level face images. 
To the best of our knowledge, this is one of the first proposal to generate intensity images from depth data for the head pose estimation task with an \textit{adversarial} approach. Moreover, we evaluate and check the overall quality of the computed face images and results confirm their high quality and accuracy.\\
%
%
Extensive experiments have been carried out and results show that the \textit{POSEidon$^+$}, equipped with the improved version of the \textit{Face-from-Depth} architecture, achieves significant improvements in the head pose estimation task. 
Besides, we show that is possible to obtain competitive results exploiting a CNN trained on gray-level faces and tested on generated ones.

\section{Related Work} \label{sec:related}
The complete framework proposed in this paper merges together several modern aspects of computer vision. Among the others, the detection, localization, and pose estimation of the head and the shoulders on depth images have been included. 
In the following, we describe the state of the art of each mentioned topic, including the \textit{Domain Translation} research area related to the \textit{Face-from-Depth} module. \\

\noindent \textbf{Head Pose.}
Head pose estimation approaches can rely on different input types: RGB images, depth maps, or both. For this reason, in order to discuss related works, we adopt a classification based on the input data types leveraged by each method.

\textit{RGB} methods take monocular or stereo intensity images as input. In \cite{whitehill2008discriminative} a discriminative approach to frame-by-frame tracking the head pose is presented, based on the detection of the centers of both eyes, the tip of the nose and the center of the mouth. Also, \cite{vatahska2007,yang2002model,matsumoto2000} leverage well visible facial features on RGB input images, and \cite{sun2008} on 3D data. \cite{drouard2017switching} proposed to predict pose parameters from high-dimensional feature vectors, embedding a Gaussian mixture of linear inverse-regression model into a dynamic Bayesian model. However, these methods need facial (\textit{e.g.} nose and eyes) or pose-dependent features, that should be always visible: consequently, these methods fail when such features are not detected.\\
A different approach for head pose estimation involves 3D model registration techniques. Firstly, Blanz and Vetter~\cite{blanz1999} propose a technique for modeling textured 3D faces automatically generated from one or more photographs. Cao \etal \cite{cao2013} exploited a 3D regression algorithm that learns an accurate, user-specific face alignment model from an easily acquired set of training data, generated from images of the user performing a sequence of predefined facial poses and expressions. Furthermore, \cite{storer20093d} proposed a hybrid approach, which exploits the flexibility of a generative 3D facial model in a combination with a fitting algorithm. However, those techniques often need a manual initialization which is indeed critical for the effectiveness of the method.\\
A first attempt to use deep learning techniques combined with the regression task in head the pose estimation problem has been performed by Ahn \etal \cite{ahn2014}, through a CNN trained on RGB input images. Also, \cite{osadchy2007synergistic} exploits a CNN by mapping images of faces on a low dimensional manifold parameterized by pose. In \cite{xu2017joint} a framework to jointly estimate the head pose and the face alignment using global and local CNN features has been presented while a hybrid approach based on CNN and Gaussian mixture was proposed in \cite{lathuiliere2017deep} and \cite{khan2017head}. With deep learning-based approaches, synthetic datasets were often used to train CNNs, that generally require a huge amount of data \cite{liu3d2016}. \\
Additionally, a bunch of methods regard head pose estimation as an \textit{optimization} problem: in \cite{bar2012} a multi-template, \textit{Iterative Closest Point} (ICP) \cite{low2004} based gaze tracking system is introduced. Besides, other works use linear or nonlinear regression with extremely low-resolution images \cite{chen2016}. HOG features and a Gaussian locally-linear mapping model were used in \cite{drouard2015} and, finally, recent works produce head pose estimations performing a face alignment task \cite{zhu2015} using CNNs.\\
In general, RGB based methods are highly sensitive to illumination, partial occlusions and bad image quality \cite{Trivedi2009}.

%

\textit{Depth} methods, on the other hand, exploit only range data to perform the pose estimation task. A first attempt to localize accurate nose locations from depth maps in order to perform head tracking and pose estimation was done in~\cite{malassiotis2005}. Consequently, \cite{breitenstein2008} used geometric features to identify nose candidates to produce the final pose estimation. A more robust approach was done in \cite{fanelli2011,fanelli2011dagm,fanelli2013}, where a Random Regression Forest \cite{liaw2002classification} algorithm is exploited for both head detection and pose estimation purposes. Furthermore, in \cite{papazov2015} facial point clouds were matched with pose candidates, through a novel triangular surface patch descriptor.\\
As previously stated for RGB methods, those techniques require facial attributes, thus are prone to errors when such features are not detected.\\
Remaining depth methods regard the head pose estimation task as an \textit{optimization} problem: \cite{padeleris2012} used the \textit{Particle Swarm Optimization} (PSO) \cite{kennedy2011} while \cite{meyer2015} perform pose estimation by registering a morphable face model to the measured depth data combining PSO and ICP techniques. Furthermore, \cite{kondori2011} used a least-square technique to minimize the difference between the input depth change rate and the prediction rate, to perform 3D head tracking. Finally, in \cite{sheng2017generative} a generative model is proposed, that unifies pose tracking and face model adaptation on-the-fly. \\
However, no previous method that uses depth maps as the only input exploits CNNs in an effective way. In this work we propose a method based on \cite{borghi17cvpr} which uses depth maps to produce accurate head pose predictions by leveraging CNNs.

\textit{RGB-D} methods combine together RGB images and depth maps. A first effort to leverage both data was done in \cite{Seemann2004}, where a Neural Network is exploited to perform head pose predictions. HOG features \cite{dalal2005histograms} were extracted from RGB and depth images in \cite{yang2012,saeed2015}, then a \textit{Multi Layer Perceptron} and a linear SVM \cite{hearst1998support} were used for feature classification, respectively. In \cite{kaymak2012exploiting} Random Forests and tensor regression algorithms are exploited while \cite{tulyakov2014} used a cascade of tree classifiers to tackle extreme head pose estimation task. Recently, in \cite{mukherjee2015} a multimodal CNN was proposed to estimate gaze direction: a regression approach was only approximated through a classifier with a granularity of \ang{1} and with 360 classes. As for RGB and depth methods, these appearance-based techniques are not robust enough: they still strongly depend on the detection of visible facial features. \\
Following 3D model registration techniques, \cite{baltruvsaitis2012} leverage intensity and depth data to build a 3D constrained local method for robust facial feature tracking. Furthermore, in \cite{ghiass2015,cai2010,bleiweiss2010,li2016real} a 3D morphable model is fitted, using both RGB and depth data to predict head pose. Finally, \cite{rekik2013}, based on a particle filter formalism, presents a new method for 3D face pose tracking in color images and depth data acquired by RGB-D cameras.\\
Several works based on head pose estimation, however, do not take in consideration the head localization task. \\

\noindent \textbf{Head Detection and Localization.} To propose a complete head pose estimation framework,  head detection is firstly required to find the complete head or a particular point, for example the head center \cite{shotton2013}. With RGB or intensity images Viola and Jones \cite{viola2004} face detector is often exploited, \textit{e.g.} in \cite{ghiass2015,cai2010,rekik2013,baltruvsaitis2012,Seemann2004}. A different approach demands the head location to a classifier, \textit{e.g., } \cite{tulyakov2014}.
As reported in \cite{meyer2015}, these approaches suffer due to the lack of generalization capabilities of exploited models, with different acquisition devices and scene contexts.\\ 
Recently, deep learning approaches trained on huge face datasets allowed to reach impressive results \cite{Faceness17,cao2017realtime}. 
However, very few works in literature propose methods for head detection or localization using \textit{only} depth images as input. A method based on a novel head descriptor and an LDA classifier is described in \cite{chen2016exploring}. Every single pixel is classified as head or non-head, and all pixels are clustered for final head detection. In \cite{nghiem2012head} a fall detection system is proposed, in which is included a module for head detection. Heads are detected only on moving objects through a background suppression. In \cite{fanelli2011} patches extracted from depth images are used to both compute the location and the pose of the head, through a regression forest algorithm.\\

\begin{figure}[b!]
    \centering
    \includegraphics[width=0.9\columnwidth]{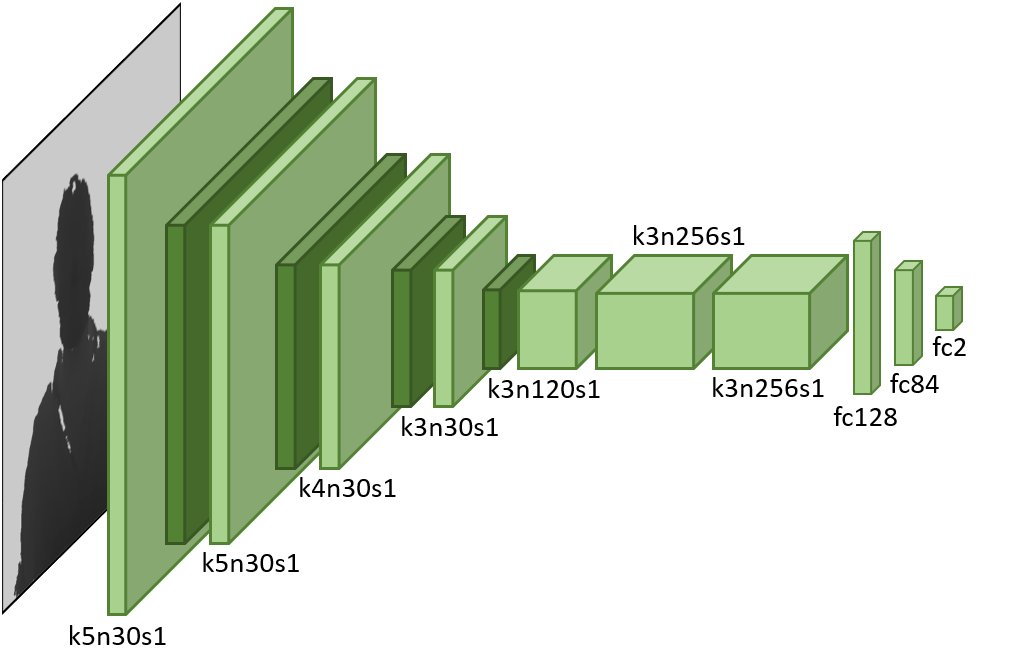}
    \caption{Architecture of the Head Localization network with corresponding kernel size (k), number of feature maps (n) and stride (s) indicated for each convolutional layer.}
    \label{fig:headlocalization}
\end{figure}

\noindent \textbf{Driver Body Pose.}
Only a limited number of works in literature tackle the problem of driver body pose estimation, focusing only on upper-body parts and taking into account automotive contexts. 
Ito \textit{et al.} \cite{ito2008}, adopting an intrusive approach, placed six marker points on the driver body to predict some typical driving operations. A 2D driver body tracking system was proposed in \cite{datta2008}, but a manual initialization of the tracking model is strictly required. In \cite{trivedi2004occupant} a thermal long-wavelength infrared video camera was used to analyze occupant position and posture. In \cite{tran2009introducing} an approach for upper body tracking system using 3D head and hands movements was developed. \\

\noindent \textbf{Domain Translation.}
Domain translation is the task of learning a parametric translation function between two domains. Recent works have addressed this problem by exploiting Conditional Generative Adversarial Networks (cGAN) \cite{mirza2014conditional} in order to learn a mapping from input to output images.\\
Isola \textit{et al.} \cite{isola2016image} demonstrated that their model, namely \textit{pix2pix}, is effective at synthesizing photos from label maps, reconstructing objects from edge maps and colorizing images. 
Moreover, Wang \textit{et al.} \cite{wang2016generative} proposed a method that acts as a rendering engine: given a synthetic scene, their \textit{Style GAN} is able to render a realistic image. 
In \cite{zhang2017improving} a cGAN is capable of translating an RGB face image to depth data.
Recently, a coupled generative adversarial networks framework has been proposed \cite{liu2016coupled}, to generate pairs of corresponding images in two different domains.
In our preliminary work \cite{borghi17cvpr}, we proposed one of the first approach, based on a traditional CNN with common aspects with respect to autoencoders \cite{masci2011stacked} and Fully Convolutional Networks \cite{long2015fully}, that was trained to compute the appearance of a face using the corresponding depth information.

\begin{figure}[t!]
    \centering
    \includegraphics[width=0.9\columnwidth]{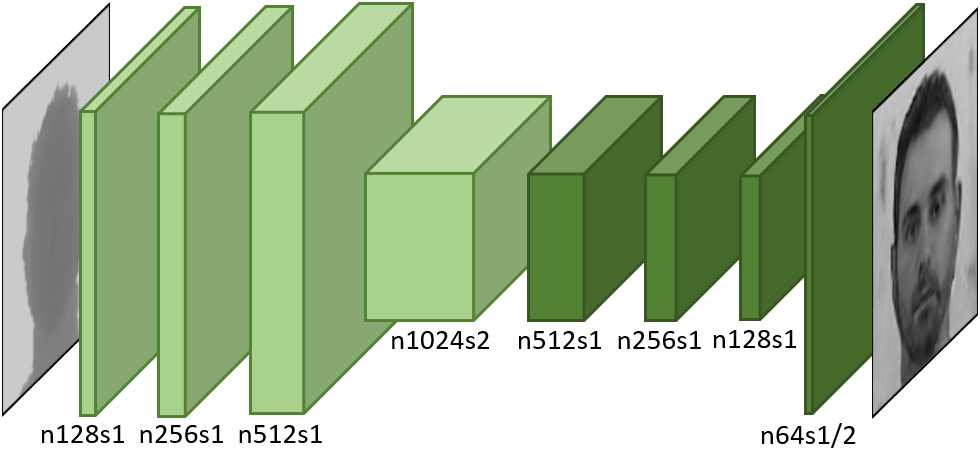}
    \caption{Architecture of the \textit{Face-from-Depth} network.}
    \label{fig:gan}
\end{figure}


\section{The POSEidon$^+$ framework}
\label{sec:architecture}
An overview of the \textit{POSEidon$^+$} framework is depicted in Figure \ref{fig:general}. The final goal is the estimation of the pose of the driver's head and shoulders, defined as the mass center position and the corresponding orientation relative to the reference frame of the acquisition device \cite{Trivedi2009}. 
The orientation is provided using three rotation angles \textit{pitch}, \textit{roll} and \textit{yaw}. 
\textit{POSEidon$^+$} directly processes the stream of depth frames captured in real-time by a commercial sensor. 
Position and size of the foremost head are estimated by a head localization module based on a regressive CNN (Sect. \ref{sec:headlocalization}). The output provided is used to crop the input frames around the head and the shoulder bounding boxes, depending on the further pipeline type. Frames cropped around the head are fed to the head pose estimation block, while the others are exploited to estimate the shoulders pose.\\
The core components of the system are the \textit{Face-from-Depth} network (Sect.~\ref{sec:Face-from-Depth}), and \textit{POSEidon$^+$} (Sect.~\ref{sec:poseidon}), the network which gives the name to the whole framework. Its trident shape is due to the three included CNNs, each working on a specific source: depth, gray level (the output of   \textit{FfD}) and \textit{Motion Images} data. The first one -- \textit{i.e.}, the CNN directly connected to the input depth data -- plays the main role on the pose estimation, while the others cooperate to reduce the estimation error.

\subsection{Head Localization} \label{sec:headlocalization}
In this step, we defined and trained a network for head localization, relying on the main assumption that a single person is in the foreground. The image coordinates $(x_H,y_H)$ of the head center are the network outputs, or rather, the center mass position of all head points in the frame~\cite{shotton2013}.\\
Details on the deep architecture adopted are reported in Figure \ref{fig:headlocalization}. A limited depth and small-sized filters have been chosen to meet real-time constraints while keeping satisfactory performance. For this reason, input images are firstly resized to $160 \times 132$ pixels. A max-pooling layer ($2 \times 2$) is run after each of the first four convolutional layers, while a dropout regularization ($\sigma=0.5$) is exploited in fully connected layers. The hyperbolic tangent activation (\textit{tanh}) function is adopted, in order to map continuous output values to a predefined range $[-\infty, +\infty] \rightarrow [-1, +1]$. The network has been trained by \textit{Stochastic Gradient Descent} (SGD) \cite{krizhevsky2012} and the $L_2$ loss function.\\ 
Given the head position $(x_H,y_H)$ in the frame, a dynamic size algorithm provides the head bounding box with center $(x_H, y_H)$, width $w_H$ and height $h_H$, around which the frames are cropped:
\begin{equation}
w_H=\frac{f_x \cdot R_x}{D}, \quad h_H=\frac{f_y \cdot R_y}{D}
\label{eq:headBB}
\end{equation}
\noindent where $f_x, f_y$ are the horizontal and the vertical focal lengths in pixels of the acquisition device, respectively. $R_x, R_y$ are the average width and height of a face (for head pose task $R_x=R_y=320$) and $D$ is the distance between the head center and the acquisition device, computed averaging the depth values around the head center. \\
Some examples of bounding boxes estimated by the network are superimposed in the frames of Figure~\ref{fig:demofinale}.

\begin{figure}[b!]
    \centering
    \includegraphics[width=0.80\columnwidth]{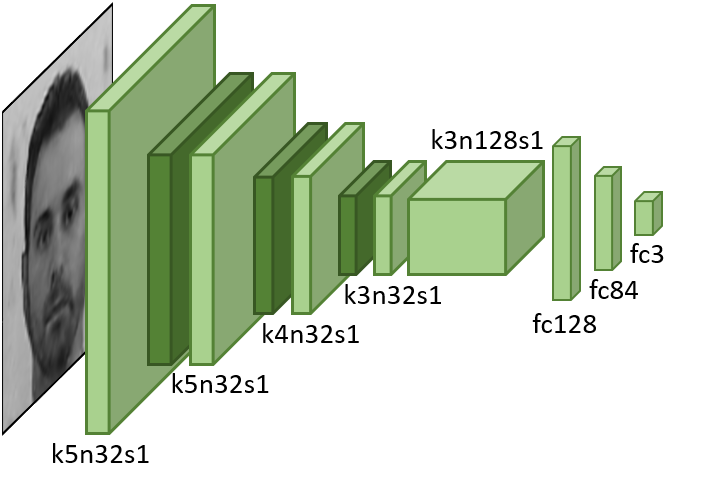}
    \caption{Architecture of the Head and Shoulder Pose Estimation networks.
    }
    \label{fig:networkpose}
\end{figure}

\section{Face-from-Depth}
\label{sec:Face-from-Depth}
 Due to illumination issues, the appearance of the face is not always available if acquired with a RGB camera, \textit{e.g.} inside a vehicle at night. On the contrary, depth maps are generally invariant to illumination conditions but lack of texture details. \\
We aim to investigate if it is possible to imagine the appearance of a face given the corresponding depth data. 
Metaphorically, we ask the model to mimic the behavior of a blind person when he tries to figure out the appearance of a friend through the touch. 

\begin{table*}[th!]
\caption{Head Pose Estimation Results on \textit{Biwi}. To allow fair comparisons with state of the art methods, \textbf{POSEidon$^+$} has been evaluated using different evaluation protocols.}
\centering
\small
\begin{tabular}{cccccccc}
\hline
\multirow{2}{*}{\textbf{Validation Procedure}} 	&\multirow{2}{*}{\textbf{Year}} &\multicolumn{2}{c}{\textbf{Data}}  &\multicolumn{3}{c}{\textbf{Head}}     &\multirow{2}{*}{\textbf{Avg}}     \\  
								&		&Depth		&RGB	 	&Pitch	&Roll		&Yaw		&		\\ \hline \multicolumn{8}{c}{}\\[-0.6em]
\textsc{All sequences used as test set } & & & & & & &\\ \hline
Padeleris \cite{padeleris2012}	&2012	&$\surd$	&			&6.6   & 6.7	& 11.1	&8.1 	 \\ \hline
Rekik \cite{rekik2013}			&2013	&$\surd$	&$\surd$	&4.3   & 5.2	& 5.1   &4.9  \\ \hline
Martin \cite{Martin2014}      	&2014   &$\surd$	&			& 2.5 	& 2.6 	  				& 3.6		 &2.9 \\ \hline
Papazov \cite{papazov2015}  	&2015  	&$\surd$	&			& 2.5 $\pm$ 7.4 		   & 3.8 $\pm$ 16.0			& 3.0 $\pm$ 9.6 	&4.0 $\pm$ 11.0  \\ \hline
Meyer \cite{meyer2015}  		&2015  	&$\surd$	&			& 2.4 		   			   & 2.1			& 2.1 	 &2.2 \\ \hline
Li \cite{li2016real}  			&2016  	&$\surd$	&$\surd$	& 1.7		   			   & 3.2			& 2.2 	 &2.4 \\ \hline
Sheng \cite{sheng2017generative}&2017  	&$\surd$	&			& 2.0		   			   & 1.9			& 2.3 	 &2.1 \\ \hline \multicolumn{8}{r}{}\\[-0.5em]
\textsc{Leave One Out (LOO) } & & & & & & &
 \\ \hline
Drouard \cite{drouard2015}  	    &2015  	&			&$\surd$	& 5.9 $\pm$ 4.8 		   & 4.7 $\pm$ 4.6			& 4.9 $\pm$ 4.1 	 &5.2 $\pm$ 4.5 \\ \hline
Drouard \cite{drouard2017switching} &2017  	&	&$\surd$			& 10.0 $\pm$ 8.7	   & 8.4 $\pm$ 8.0  & 8.6 $\pm$ 7.2 	 &9.0 $\pm$ 7.9 \\ \hline 
\textbf{POSEidon$^+$}   			&2017  	&$\surd$	&			&  \textbf{2.4} $\pm$ \textbf{1.3}    &  \textbf{2.6} $\pm$ \textbf{1.5}    &  \textbf{2.9} $\pm$ \textbf{1.5}    &\textbf{2.6} $\pm$ \textbf{1.4}  \\ \hline \multicolumn{8}{c}{}\\[-0.6em]  

\textsc{k4-fold subject cross validation}   	& 	&	&			&   &		&   	&  \\ \hline
Fanelli \cite{fanelli2013}       	&2011   &$\surd$	&			& 3.5 $\pm$ 5.8			   & 5.4 $\pm$ 6.0			& 3.8 $\pm$ 6.5		&- $\pm$ -   \\ \hline 
\textbf{POSEidon$^+$}   			&2017  	&$\surd$	&			&  \textbf{2.8} $\pm$ \textbf{1.7}    &  \textbf{2.9} $\pm$ \textbf{2.1}    &  \textbf{3.6} $\pm$ \textbf{2.5}    &\textbf{3.1} $\pm$ \textbf{2.1}  \\ \hline \multicolumn{8}{c}{}\\[-0.6em]

\textsc{k5-fold subject cross validation}   	& 	&	&			&   &		&   	&  \\ \hline
Fanelli \cite{fanelli2011dagm}  	&2011   &$\surd$	&			& 8.5 $\pm$ 9.9			   & 7.9 $\pm$ 8.3			& 8.9 $\pm$ 13.0	&8.43 $\pm$ 10.4   \\ \hline
\textbf{POSEidon$^+$}   			&2017  	&$\surd$	&			& \textbf{2.8} $\pm$ \textbf{1.8}  & \textbf{2.8} $\pm$ \textbf{2.2}  	   			& \textbf{3.6} $\pm$ \textbf{2.2}  	&\textbf{3.0} $\pm$ \textbf{2.1}  \\ \hline \multicolumn{8}{c}{}\\[-0.6em]

\textsc{k8-fold subject cross validation}   	& 	&	&			&   &		&   	&  \\ \hline
Lathuiliere \cite{lathuiliere2017deep}&2017  	&	&$\surd$			& 4.7		   	   & 3.1			& \textbf{3.1} 	 &3.6 \\ \hline 
\textbf{POSEidon$^+$}   			&2017  	&$\surd$	&			&  \textbf{2.8} $\pm$ \textbf{1.9}    &  \textbf{2.8} $\pm$ \textbf{1.8}    & 3.3 $\pm$ 2.0    &\textbf{3.0} $\pm$ \textbf{1.9}  \\ \hline \multicolumn{8}{c}{}\\[-0.6em]  
\textsc{Fixed train and test splits} & & & & & & &\\ \hline
Yang \cite{yang2012}			&2012	&$\surd$	&$\surd$	& 9.1 $\pm$ 7.4			   & 7.4 $\pm$ 4.9			& 8.9 $\pm$ 8.3	&8.5 $\pm$ 6.9 	 \\ \hline
Baltrusaitis  \cite{baltruvsaitis2012}  	&2012    	&$\surd$ &$\surd$	& 5.1				& 11.3					& 6.3	&7.6 \\ \hline
Kaymak \cite{kaymak2012exploiting}	&2013	&$\surd$	&$\surd$			&7.4   &6.6	& 5.0   &6.3  \\ \hline
Wang \cite{wang2013head}		&2013	&$\surd$	&$\surd$	&8.5 $\pm$ 14.3  & 7.4 $\pm$ 10.8	& 8.8 $\pm$ 14.3   & 8.2 $\pm$ 12.0 \\ \hline
Ahn \cite{ahn2014}      		    &2014   & 			&$\surd$	& 3.4 $\pm$ 2.9			   & 2.6 $\pm$ 2.5			& 2.8 $\pm$ 2.4		&2.9 $\pm$ 2.6  \\ \hline
Saeed  \cite{saeed2015}      	&2015   &$\surd$ 	&$\surd$	& 5.0 $\pm$ 5.8			   & 4.3 $\pm$ 4.6			& 3.9 $\pm$ 4.2		&4.4 $\pm$ 4.9  \\ \hline
Liu \cite{liu3d2016}  		&2016  	&			&$\surd$	& 6.0 $\pm$ 5.8 		   & 5.7 $\pm$ 7.3			& 6.1 $\pm$ 5.2 	 &5.9 $\pm$ 6.1 \\ \hline
POSEidon \cite{borghi17cvpr}   	    &2017  	&$\surd$	&			& 1.6 $\pm$ 1.7  & 1.8 $\pm$ 1.8		& 1.7 $\pm$ 1.5  	&1.7 $\pm$ 1.7  \\ \hline
\textbf{POSEidon$^+$}   			&2017  	&$\surd$	&			& \textbf{1.6} $\pm$ \textbf{1.3}  & \textbf{1.7} $\pm$ \textbf{1.7}  	   			& \textbf{1.7} $\pm$ \textbf{1.3}  	&\textbf{1.6} $\pm$ \textbf{1.4}  \\ \hline

\end{tabular}
\label{tab:resBiwiPose}
\end{table*}

\subsection{Deterministic Conditional GAN}
The \textit{Face-From-Depth} network exploits the \textit{Deterministic Conditional GAN} (det-cGAN) paradigm \cite{isola2016image} and it is obtained as a generative network $G$ capable of estimating a gray-level image $I^E$ of a face from the corresponding depth representation $I^D$. 
The generator $G$ is trained to produce outputs as much indistinguishable as possible from \textit{real} images $I$ by an adversarially trained discriminator $D$, which is expressly trained to distinguish the \textit{real} images from the \textit{fake} ones produced by the generator. 
Differently from a traditional GAN \cite{goodfellow2014generative,radford2015unsupervised}, the Generator network of a det-cGAN takes an image as input (to be \textit{Conditional}) and not a random noise vector (to be \textit{Deterministic}). As a result, a det-cGAN learns a mapping from observed images $x$ to output images $y$: $G:x\rightarrow y$.

The objective of a det-cGAN can be expressed as follows:
\begin{multline}
\label{eq:ldetcgan}
	 L_{det-cGAN}(G,D)=\mathbb{E}_{I \sim p_{data}(I)}[\log{D(I)}] \\
	+ \mathbb{E}_{I^D \sim p_{data}(I^D)}[\log({1 - D(G(I^D))})]
\end{multline}
\noindent where $\log{D(I)}$ represents the log probability that $I$ is \textit{real} rather than \textit{fake} while $\log({1 - D(G(I^D))})$ is the log probability that $G(I^D)$ is \textit{fake} rather than \textit{real}.
$G$ tries to minimize the term $L_{det-cGAN}(G,D)$ of Equation~\ref{eq:ldetcgan}, against $D$ that tries to maximize it. The optimal solution is:
\begin{equation}
\label{eq:eg}
	G^*=\arg \min_{G} \max_{D}L_{det-cGAN}(G,D)
\end{equation}


\noindent As a possible drawback, the images generated by $G$ are forced to be realistic thanks to $D$, but they can be unrelated with the original input. For instance, the output could be a nice image of a head with a very different pose with respect to the input depth. 
Thus, is fundamental mixing the GAN objective with a more traditional loss, such as SSE distance
\cite{pathak2016context}. While discriminator’s job remains unchanged, the generator, in addition to fooling the discriminator, tries to emulate the ground truth output in an SSE sense. The pixel-wise SSE is calculated between downsized versions of the generated and target images, first applying an averaged pooling layer. 
We formulate the final objective as the weighted sum of a content loss and an adversarial loss as:
\begin{equation}
\label{eq:final}
	G^*=\arg \min_{G} \max_{D}L_{det-cGAN}(G,D) + \lambda L_{SSE}(G)
\end{equation}
\noindent where $\lambda$ is the weight controlling the content loss impact.

\begin{table*}[t]
\caption{Evaluation metrics computed on the reconstructed gray-level face images with
\textit{Biwi} and \textit{Pandora} datasets. Starting from the left, $L_1$ and $L_2$ distances are reported, then the absolute and the squared differences, the root-mean-square error and, finally, the percentage of pixels  under a certain threshold. Further details about metrics are reported in \cite{NIPS2014}.}
\centering
\small
\begin{tabular}{c|c||cc|cc|ccc|ccc}
\hline
\multirow{2}{*}{\textbf{Dataset}} & \multirow{2}{*}{\textbf{Method}}  &\multicolumn{2}{c|}{\textbf{Norm} $\downarrow$} &\multicolumn{2}{c|}{\textbf{Difference} $\downarrow$}  &\multicolumn{3}{c|}{\textbf{RMSE} $\downarrow$} &\multicolumn{3}{c}{\textbf{Threshold} $\uparrow$} \\
\cline{3-12}
& &$L_1$ &$L_2$ &Abs  &{\footnotesize Squared}  &linear &log &scale-inv &1.25 &2.5 &3.75 \\ \hline
\hline
\multirow{2}{*}{Biwi} &  FfD \cite{borghi17cvpr} &33.35 &2586 &0.454 &24.07 &40.55 &\textbf{0.489} & \textbf{0.445} &0.507 &\textbf{0.806} & \textbf{0.878} \\ \cline{2-12}
& \textbf{FfD} 		&\textbf{24.44} &\textbf{2230} &\textbf{0.388} &\textbf{19.81} &\textbf{35.50} &0.653 & 0.610 &\textbf{0.615} &0.764 &0.840 \\ 
\hline \hline
\multirow{5}{*}{Pandora} & FfD \cite{borghi17cvpr}	&41.36 &3226 &0.705 &46.00 &50.77 &0.603 &\textbf{0.485} &0.263 &0.725 &0.819 \\ \cline{2-12}
 & pix2pix \cite{isola2016image}	&19.37 &1909 &\textbf{0.468} &24.07 &30.80 &0.568 &0.539 &0.583 &0.722 &0.813 \\ \cline{2-12}
& AVSS \cite{matteo2017generative}	&23.93 &2226 &0.629 &34.49 &35.46 &0.658 &0.579 &0.541 &0.675 &0.764 \\ \cline{2-12}
& FfD + U-Net	&23.75 &2123 &0.653 &34.96 &33.89 &0.639 &0.553 &0.555 &0.689 &0.775 \\ \cline{2-12}
& \textbf{FfD}	&\textbf{18.21} &\textbf{1808} &0.469 &\textbf{22.90} &\textbf{28.90} &\textbf{0.556} &0.501 &\textbf{0.605} &\textbf{0.743} &\textbf{0.828} \\ 
\hline
%
\end{tabular}
\label{tab:metrics}
\end{table*}

\subsection{Network Architecture} \label{sec:gan_architecture}
We propose to modify the classic hourglass generator architecture, performing a limited number of upsampling and downsampling operations. 
As shown in the following experimental section, the \textit{U-Net} architecture \cite{ronneberger2015u} can be adopted in order to shuttle low-level information between input and output directly across the network \cite{isola2016image}, but it is less convenient in our case.\\
Following the main architecture guidelines for stable Deep Convolutional GANs by Radford \textit{et al.} \cite{radford2015unsupervised}, we instead adopt the architecture illustrated in Figure \ref{fig:gan} for the Generator. 
Specifically, in the encoder part, we use three convolutional layers followed by a strided convolutional layer (with stride 2) to halve the image resolution.\\ 
The decoding stack uses three convolutional layers followed by a transposed convolutional layer (also referred as fractionally strided convolutional layers) with stride $1/2$ to double the resolution, and a final convolution. The number of filters follows a power of 2 pattern, from 128 to 1024 in the encoder and from 512 to 64 in the decoder. 
\textit{Leaky ReLU} is used as activation function in the encoding phase while \textit{ReLU} is used in the decoding phase. \\
We adopt \textit{batch normalization} before each activation (except for the last layer) and a kernel size $5 \times 5$ for each convolution.\\ 
The discriminator architecture complies with the generator’s encoder in terms of activation and number of filters, but contains only strided convolutional layers (with stride 2) to halve the image resolution each time the number of filters is doubled. The network then outputs one \textit{sigmoid} activation.
In the discriminator, we use batch normalization before every Leaky ReLU activation, except for the first layer.

\subsection{Training details}
We trained the det-cGAN with depth images and simultaneously providing the network with the original gray-level images associated with the depth data in order to compute the $L_{SSE}$. To optimize the network we adopted the standard approach from Goodfellow \textit{et al.} \cite{goodfellow2014generative} and alternate the gradient descent updates between the generator and the discriminator with $K=1$. We used mini-batch SGD applying the \textit{Adam} solver \cite{kingma2014adam} with $\beta_1 = 0.5$ and batch size of $64$. We set $\lambda=10^{-1}$ in Equation~\ref{eq:final} for the experiments. Moreover, to encourage the discriminator to estimate soft probabilities rather than to extrapolate extremely confident classifications, we used a technique called \textit{one-sided label smoothing} \cite{salimans2016improved} where the target for the real examples are replaced with a value slightly less than $1$, such as $0.9$. This solution prevents the discriminator to produce extremely confident predictions that could unbalance the adversarial learning. 

\section{Pose Estimation from depth}\label{sec:poseidon}
\subsection{POSEidon$^+$ network} \label{sec:poseidonpose}
The \textit{POSEidon$^+$} network is a fusion of three CNNs and has been developed to perform a regression on the 3D pose angles. As a result, continuous Euler values -- corresponding to the \textit{yaw}, \textit{pitch} and \textit{roll} angles -- are estimated (right part of Fig. \ref{fig:general}). 
The three \textit{POSEidon$^+$} components have the same shallow architecture based on 5 convolutional layers with kernel size of $5 \times 5$, $4 \times 4$ and $3 \times 3$ and a $2 \times 2$ max-pooling is conducted only on the first three layers due to the limited size of the input ($64 \times 64$). 
The first four convolutional layers have 32 filters each, the last one has 128 filters.
\textit{tanh} is exploited as activation function; we are aware that \textit{ReLU} \cite{nair2010rectified} converges faster, but better performance in term of accuracy prediction are achieved.\\ 
The three networks are fed with different input data types: the first one, directly takes as input the head-cropped depth images; the second one is connected to the \textit{Face-from-Depth} output and the last one operates on Motion Images, obtained applying the standard \textit{Farneback} algorithm \cite{farneback2001} on pairs of consecutive depth frames. 
The presence of depth discontinuities around the nose and the eyes generates specific motion patterns which are related to the head pose. Motion Images, thus, provide useful information for the estimation of the pose of a moving head. Frames with motionless heads are very rare in real videos. However, in those cases the common image compression creates artifacts around the face landmarks which allow the estimation of the head pose. 

A fusion step combines the contributions of the three above described networks. The last fully connected layer of each component is removed in order to provide the following layers with more data and not only the estimated angles. As a results, the output of the whole \textit{POSEidon$^+$}  network is not only a weighted mean of the three component outputs, but a more complex combination.
Different fusion approaches that have been proposed by Park \etal \cite{park2016} are investigated. Given two feature maps $x^a, x^b$ with a certain width $w$ and height $h$, for every feature channel $d^x_a, d^x_b$ and $y \in R^{w \times h \times d}$:
\begin{itemize}
\item \textbf{Multiplication}: computes the element-wise product of two feature maps, as
$ y^{mul}=x^a \circ x^b, d^y = d^x_a = d^x_b $
\item \textbf{Concatenation}: stacks two features maps, without any blend
$y^{cat} = [x^a | x^b], d^y = d^x_a + d^x_b$
\item \textbf{Convolution}: stacks and convolves feature maps with a filter $k$ of size $1 \times 1 \times (d^x_a + d^x_b)/2$ and $\beta$ as bias term, 
$ y^{conv} = y^{cat} \ast k + \beta, \quad d^y = (d^x_a+d^x_b) / 2 $
\end{itemize}

\begin{figure*}[t!]
\centering
\subfigure[]{\includegraphics[width=0.607\linewidth]{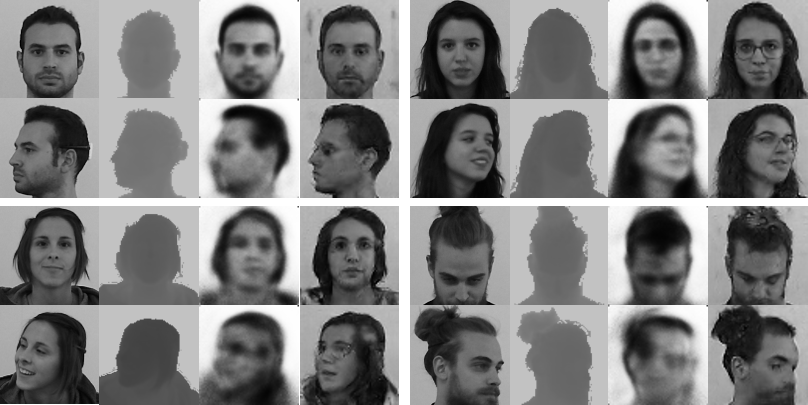}} 
\quad
\subfigure[]{\includegraphics[width=0.3\linewidth]{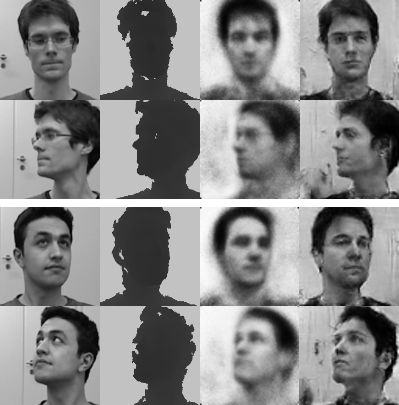}}
\subfigure[]{\includegraphics[width=0.607\linewidth]{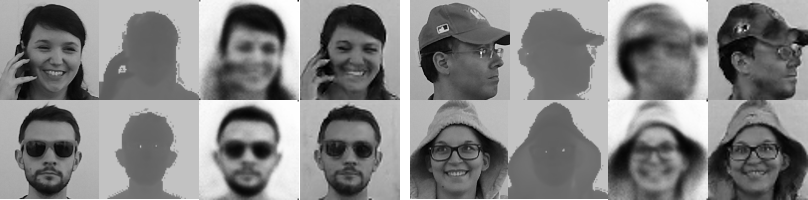}}
\quad
\subfigure[]{\includegraphics[width=0.3\linewidth]{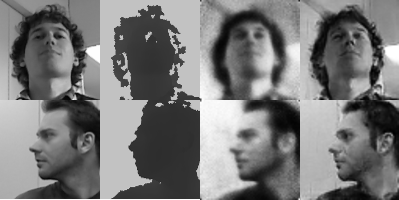}}
\caption{Test (a) and train (c) images on \textit{Pandora} dataset, test (b) and train (d) images on \textit{Biwi} dataset. For each block, gray-level images and then the corresponding depth faces are depicted in the first columns; face images taken from the method described in \cite{borghi17cvpr} are reported in the third column; finally, the output of the \textit{Face-from-Depth} network proposed in this paper is depicted in the last column. }
\label{fig:facefromdepth} 
\end{figure*}
%
The final \textit{POSEidon$^+$} framework exploits a combination of two fusing methods, in particular, a convolution followed by a concatenation. After the fusion step, three fully connected layers composed of 128, 84 and 3 activations respectively and two dropout regularization ($\sigma=0.5$) complete the architecture.
\textit{POSEidon$^+$} is trained with a double-step procedure. First, each individual network described above is trained with the following $L_2^w$ weighted loss:
\begin{equation}
L_2^w = \sum_{i=1}^3 \big \lVert w_i \cdot \left(y_i - f(x_i)\right) \big \rVert _2
\label{eq:l2_pesata}
\end{equation}
where $ w_i \in [0.2, 0.35, 0.45]$. This weight distribution gives more importance to the yaw angle, which is preponderant in the selected automotive context. During the individual training step, the last fully connected layer of each network is preserved, then is removed to perform the second training phase. Holding the weights learned for the trident components, the new training phase is carried out on the last three fully connected layers of \textit{POSEidon$^+$} only, with the loss function $L^w_2$ reported in Equation \ref{eq:l2_pesata}. 
In all training steps, the SGD optimizer \cite{krizhevsky2012} is exploited, the learning rate is set initially to $10^{-1}$ and then is reduced by a factor 2 every 15 epochs.

\subsection{Shoulder Pose Estimation} \label{sec:shoulderpose}
The framework is completed with an additional network for the estimation of the shoulder pose. We employ the same architecture adopted for the head (Sect. \ref{sec:poseidonpose}), performing regression on the  three pose angles.\\
Starting from the head center position (Sect. \ref{sec:headlocalization}), the depth input images are cropped around the driver neck, using a bounding box $\{x_S, y_S, w_S, h_S\}$ with center $(x_S = x_H, y_S = y_H - (h_H/4))$, and width and height obtained as described in Equation \ref{eq:headBB}, but with different values of $R_x, R_y$ to produce a rectangular crop: these values are tested and discussed in Section \ref{sec:results}.
The network is trained with SGD optimizer \cite{krizhevsky2012}, using the weighted $L_2^w$ loss function described above (see Eq.~\ref{eq:l2_pesata}). As usual, hyperbolic tangent is exploited as activation function.

\section{Experimental Results}
\label{sec:results}
\subsection{Datasets}
Network training and testing phases have been done exploiting two publicly available datasets, namely \textit{Biwi Kinect Head Pose} and \textit{ICT-3DHP}. In addition, we collected a new dataset, called \textit{Pandora}, which also contains shoulder pose annotations. 
Data augmentation techniques are employed to enlarge the training set, in order to achieve space invariance and avoid overfitting \cite{krizhevsky2012}.\\
Random translations on vertical, horizontal and diagonal directions, jittering, zoom-in and zoom-out transformation of the original images have been exploited. Percentile-based contrast stretching, normalization and scaling of the input images are also applied to produce zero mean and unit variance data. \\
Other datasets for head pose estimation and related tasks have been collected in last decades \cite{Masi,gourier2004estimating,Nuevo2010,yuen2016looking,martin2016vision}, but in most cases there are some not desirable features, for instance, no depth or 3D data, no continuous ground truth annotations and not enough data for deep learning techniques.\\
Follows a detailed description of the three adopted datasets. 

\subsubsection{Biwi Kinect Head Pose dataset}
Fanelli \textit{et al.} \cite{fanelli2011} introduced this dataset in 2013. It is acquired with the \textit{Microsoft Kinect} sensor, i.e., a structured IR light device. It contains about 15k frames, with RGB ($640 \times 480$) and depth maps ($640 \times 480$). Twenty subjects have been involved in the recordings: four of them were recorded twice, for a total of 24 sequences. The ground truth of yaw, pitch and roll angles is reported together with the head center and the calibration matrix. The original paper does not report the adopted split between training and testing sets; fair comparisons are thus not guaranteed. To avoid this, we clearly report the adopted split in the following.

\subsubsection{ICT-3DHP dataset}
\textit{ICT-3DHP} dataset has been introduced by Baltrusaitis \textit{et al.} in 2012 \cite{baltruvsaitis2012}. It has been collected using a \textit{Microsoft Kinect} sensor and contains RGB images and depth maps of about 14k frames, divided into 10 sequences. The image resolution is $640 \times 480$ pixels. An additional hardware sensor (\textit{Polhemus Fastrack}) is exploited to generate the ground truth annotation. The device is placed on a white cap worn by each subject, visible in both RGB and depth frames. The presence of a few subjects and the limited number of frames make this dataset unsuitable for training deep learning approaches.

\subsubsection{Pandora dataset} \label{sec:dataset}
In addition to publicly available datasets, we have also collected and used a new challenging dataset, called \textit{Pandora}. 
It has been specifically created for head center localization, head pose and shoulder pose estimation in automotive contexts (See Fig.\ref{fig:pandora}). A frontal and fixed device acquires the upper body part of the subjects, simulating the point of view of a camera placed inside the dashboard. The subjects mainly perform driving-like actions, such as holding the steering wheel, looking to the rear-view or lateral mirrors, shifting gears and so on. \textit{Pandora} contains 110 annotated sequences of 10 male and 12 female actors. Each subject has been recorded five times. 
\textit{Pandora} is the first publicly available dataset which combines the following features:
\begin{itemize}
\item \textbf{Shoulder angles}: in addition to the head pose annotation, \textit{Pandora} contains the ground truth data of the shoulder pose expressed as yaw, pitch, and roll. 
\item \textbf{Wide angle ranges}: subjects perform wide head ($\pm \ang{70}$ roll, $\pm \ang{100}$ pitch and $\pm \ang{125}$ yaw) and shoulder ($\pm \ang{70}$ roll, $\pm \ang{60}$ pitch and $\pm \ang{60}$ yaw) movements. For each subject, two sequences are performed with constrained movements, changing the yaw, pitch and roll angles separately, while three additional sequences are completely unconstrained.
\item \textbf{Challenging camouflage}: garments, as well as various objects are worn or used by the subjects to create head and/or shoulder occlusions. For example, people wear prescription glasses, sunglasses, scarves, caps, and manipulate smart-phones, tablets or plastic bottles. 
\item \textbf{Deep-learning oriented}: the dataset contains more than 250k full resolution RGB ($1920 \times 1080$) and depth images ($512 \times 424$) with the corresponding annotation.
\item \textbf{Time-of-Flight (ToF) data}: a \textit{Microsoft Kinect One} device is used to acquire depth data, with a better quality than other datasets created with the first \textit{Kinect} version~\cite{sarbolandi2015}. 
\end{itemize}

Each frame of the dataset is composed of an RGB appearance image, the corresponding depth map, and the 3D coordinates of the skeleton joints corresponding to the upper body part, including the head center and the shoulder positions. For convenience's sake, the 2D coordinates of the joints on both color and depth frames are provided as well as the head and shoulder pose angles with respect to the camera reference frame. Shoulder angles are obtained through the conversion to Euler angles of a corresponding rotation matrix, obtained from a user-centered system~\cite{papadopoulos2014} and defined by the following unit vectors $(N_1,N_2,N_3)$:
\begin{equation}
\begin{array}{cc}
N_1=\frac{p_{RS}-p_{LS}}{\| p_{RS}-p_{LS} \|} & U=\frac{p_{RS}-p_{SB}}{\|p_{RS}-p_{SB}\|}\\ \\
N_3=\frac{N_1 \times U}{\| N_1 \times U \|} & N_2=N_{1} \times N_{3}
\end{array}
\label{eq:refframe}
\end{equation}
where $p_{LS}$, $p_{RS}$ and $p_{SB}$ are the 3D coordinates of the left shoulder, right shoulder and spine base joints.
The annotation of the head pose angles has been collected using a wearable \textit{Inertial Measurement Unit} (IMU) sensor. 
To avoid distracting artifacts on both color and depth images, the sensor has been placed in a non-visible position,  \textit{i.e.}, on the rear of the subject's head. The IMU sensor has been calibrated and aligned at the beginning of each sequence, assuring the reliability of the provided angles. The dataset is publicly available (\url{http://aimagelab.ing.unimore.it/pandora/}).

\begin{table}[b]
\caption{Results obtained on \textit{Pandora} dataset with head pose network trained on gray level images and tested with the original gray-level and reconstructed ones. }
\centering
\small
\begin{tabular}{ccccc}
\hline
\multirow{2}{*}{\textbf{Testing input}}  &\multicolumn{3}{c}{\textbf{Head}} &\multirow{2}{*}{\textbf{Acc.}}   \\
 						&Pitch		&Roll			&Yaw    &				\\ \hline \hline
\textbf{gray-level}	&\textbf{7.1} $ \pm$ \textbf{5.6}		&\textbf{5.6} $\pm$ \textbf{5.8}    &\textbf{9.0} $\pm$ \textbf{10.9}	&\textbf{0.613}  \\ \hline \hline                        
pix2pix \cite{isola2016image}		&7.9 $ \pm$ 8.0		&5.9 $\pm$ 6.3    &12.8 $\pm$ 21.4	&0.581  \\ \hline
AVSS \cite{matteo2017generative}		&8.9 $ \pm$ 8.5		&6.2 $\pm$ 6.4    &13.4 $\pm$ 20.4	&0.543  \\ \hline
FfD \cite{borghi17cvpr}   		&8.5 $ \pm$ 8.9		&6.1 $\pm$ 6.2    &12.4 $\pm$ 17.3	&0.559  \\ \hline
FfD + U-Net		&8.7 $ \pm$ 8.4		&6.4 $\pm$ 6.6    &13.5 $\pm$ 19.9  &0.552  \\ \hline
\textbf{FfD} 		&\textbf{7.6} $ \pm$\textbf{ 6.9}		&\textbf{5.8} $\pm$ \textbf{6.0}    &\textbf{10.1} $\pm$\textbf{ 12.6}	&\textbf{0.613}  \\ \hline

\end{tabular}
\label{tab:testgenerated}
\end{table}

\begin{table*}[th!]
\caption{Results of the head pose estimation on \textit{Pandora} comparing different system architectures. The baseline is a single CNN working on the source depth map (Row 1). The accuracy is the percentage of correct estimations ($err<$\ang{15}). FfD: \textit{Face-from-Depth}, MI: \textit{Motion Images}.}
\centering
\small
\begin{tabular}{c cccc c ccccc}
\multicolumn{10}{c}{\textsc{Head Pose estimation error [euler angles]}}\\ \hline
\multirow{2}{*}{\textbf{\#}} &
\multicolumn{4}{c}{\textbf{Input}} &\multirow{2}{*}{\textbf{Crop}} &\multirow{2}{*}{\textbf{Fusion}} &\multicolumn{3}{c}{\textbf{Head}}   &\multirow{2}{*}{\textbf{Accuracy}} \\
& Depth	&FfD 		&MI		&Gray 	& 		&			 &Pitch					&Roll			  &Yaw		   	  	 &	\\ \hline \hline

1& $\surd$	& 			& 		&    	& 		&-			 &8.1 $\pm$ 7.1 		&6.2 $\pm$ 6.3		&11.7 $\pm$ 12.2 	 &0.553	\\ \hline
2&$\surd$ &			&		&  		&$\surd$ 		&-	 &6.5 $\pm$ 6.6 		&5.4 $\pm$ 5.1		&10.4 $\pm$ 11.8 	 &0.646	\\ \hline
3&		&$\surd$	&  		& 		&$\surd$   	&-	 &6.8 $\pm$ 6.1 		&5.8 $\pm$ 5.0		&10.1 $\pm$ 12.6 	 &0.658	\\ \hline
4&        & 			&$\surd$&  		&$\surd$	  &-			 &7.7 $\pm$ 7.5 		&5.3 $\pm$ 5.7		&10.0 $\pm$ 12.5 	 &0.609	\\ \hline
5&		&		 	& 		&$\surd$&$\surd$		&-	 &7.1 $\pm$ 6.6 		&5.6 $\pm$ 5.8		&9.0 $\pm$ 10.9 	 &0.639	\\ \hline
6&$\surd$	&$\surd$ 	&		&		&$\surd$	&concat		 &5.6 $\pm$ 5.0 		&4.9 $\pm$ 5.0		&9.7  $\pm$ 12.1	 &0.698	\\ \hline
7&$\surd$	&		 	&$\surd$&  		&$\surd$	&concat		 &6.0 $\pm$ 6.1 		&4.5 $\pm$ 4.8		&9.2 $\pm$11.5       &0.690	\\ \hline
8&$\surd$	&$\surd$	&$\surd$&		&$\surd$	&conv+concat 	&\textbf{5.6} $\pm$ \textbf{5.2}			&\textbf{4.8} $\pm$ \textbf{5.0}      &\textbf{8.2} $\pm$ \textbf{9.8}     &\textbf{0.736}	\\ \hline
\end{tabular}
\label{tab:resOurDataset}
\end{table*}

\subsection{Quantitative Results}
The proposed framework has been deeply tested using the datasets described in Section \ref{sec:dataset}. 
For evaluation with the \textit{Pandora} dataset, sequences of subjects 10, 14, 16 and 20 have been used for testing, the remaining for training and validation. 
With \textit{Biwi} dataset, test subjects are determined by the validation procedure adopted. 
Finally, we tested the system on all the sequences contained in \textit{ICT-3DHP} dataset.\\

\noindent \textbf{Domain Translation. } 
First, we check the capabilities of the  \textit{Face-from-Depth} network alone. 
Some visual examples of input, output, and ground-truth images are reported in Figure \ref{fig:facefromdepth}.\\
With this aim, we propose two types of evaluation. 
The first is based on metrics related to the reconstruction accuracy. 
Following the work of Eigen etal \cite{NIPS2014}, Table \ref{tab:metrics} reports some results. 
The system is evaluated both on \textit{Biwi} and on \textit{Pandora} datasets. \textit{FfD} network is compared with other Image-to-Image methods taken from the recent literature. In particular, we trained from scratch the deep models proposed in \cite{isola2016image,matteo2017generative} (referred here as \textit{pix2pix} and \textit{AVSS}, respectively), following procedures reported in the corresponding papers.\\
Moreover, in order to investigate how architectural choices impact the reconstruction quality of \textit{FfD}, we tested a different design.
We modified the network adding the \textit{U-Net} \cite{ronneberger2015u} skip connections between mirrored layers (cf. Sect. \ref{sec:gan_architecture}). \\
We also compared the presented approach with our preliminary version of \textit{Face-from-Depth} network \cite{borghi17cvpr}, that fuses the key aspects of \textit{encoder-decoder} \cite{masci2011stacked} and \textit{fully convolutional} \cite{long2015fully} neural networks. \\
For the sake of comparison, we report here key details about the preliminary \textit{FfD} version \cite{borghi17cvpr}. It has been trained in a single step, with input head images resized to $64 \times 64$ pixels. The activation function is the hyperbolic tangent and best training performance are reached through the self adaptive \textit{Adadelta} optimizer \cite{zeiler2012adadelta}. A particular loss function is exploited in order to highlight the central area of the image, where the face is supposed to be after the cropping step, and takes in account the distance between the reconstructed image and the corresponding gray-level ground truth:
\begin{equation}
L = \frac{1}{R\cdot C} \sum_i^R \sum_j^C \left( || y_{ij} - \bar{y}_{ij} || ^ 2 _2 \cdot w_{ij}^{\mathcal{N}} \right)
\end{equation}
\noindent where $R,C$ are the number of rows and columns of the input images, respectively. $y_{ij}, \bar{y}_{ij} \in \mathcal{R}^{ch}$ are the intensity values from ground truth ($ch=1$) and predicted appearance images. Finally, the term $w_{ij}^{\mathcal{N}}$ introduces a bivariate Gaussian prior mask. Best results have been obtained using $\mu=[\frac{R}{2},\frac{C}{2}]^T$ and $\Sigma = \mathbb{I} \cdot [ \left(R / \alpha \right)^2, \left( C / \beta \right)^2 ]^T$ with $\alpha$ and $\beta$ empirically set to $3.5, 2.5$ for squared images of $R=C=64$.
Other details about network architecture and training are reported in \cite{borghi17cvpr}.  

The second set of tests is specific to the head pose estimation task. The head pose network described in Section \ref{sec:poseidon}, trained with gray-level images taken from the \textit{Pandora} dataset, is tested on the reconstructed face images.
Since the network has been trained on real gray-level images to output the angles of the head pose, we can suppose that the more generated images are similar to the corresponding gray-level ones, the better the results are. 
The comparison is presented in Table \ref{tab:testgenerated}. In the first row, results obtained using gray-level images as testing input are reported, this is the best case and should be used as a reference baseline. Results present in the following rows confirm that our \textit{FfD} is able to generate high-quality faces, very similar to gray-level faces. Moreover, we note that the head pose network has the ability to generalize well on cross-dataset evaluations since we generally obtain a good accuracy even with different types of face images as input.\\ 
The \textit{Face-from-Depth} network has been created to this goal, even if the output is not always realistic and visually pleasant: however, the promising results confirm their positive contribution in the estimation of the head pose.\\

\begin{table}[b!]
\caption{Results for head pose estimation task on \textit{Pandora} dataset. In particular, here we compare our preliminary work \cite{borghi17cvpr} with the proposed one. In addition, we include a comparison with \textit{POSEidon$^+$} framework, in which we replace the head pose estimation network trained on reconstructed face images with the same network trained on gray-level images, here referred as \textit{POSEidon}*.}
\centering
\small
\begin{tabular}{ccccc}
\hline
\multirow{2}{*}{\textbf{Method}}  &\multicolumn{3}{c}{\textbf{Head}} &\multirow{2}{*}{\textbf{Acc.}}   \\
 						&Pitch		&Roll			&Yaw    &				\\ \hline \hline
POSEidon \cite{borghi17cvpr}	&5.7 $ \pm$ 5.6		&4.9 $\pm$ 5.1    &9.0 $\pm$ 11.9	&0.715  \\ \hline
POSEidon*					&5.6 $ \pm$ 5.8		&4.8 $\pm$ 5.0    &8.8 $\pm$ 10.9	&0.720  \\ \hline
\textbf{POSEidon$^+$}						&\textbf{5.6} $ \pm$ \textbf{5.2}		&\textbf{4.8} $\pm$\textbf{ 5.0}    &\textbf{8.2} $\pm$ \textbf{9.8}	&\textbf{0.736}  \\ \hline
\end{tabular}
\label{tab:testposeidon}
\end{table}

\noindent \textbf{Head Pose Estimation. }  
An ablation study of \textit{POSEidon$^+$} framework on \textit{Pandora} is conducted and results are reported in Table \ref{tab:resOurDataset}, providing mean and standard deviation of the estimation errors obtained on each angle and for each system configuration. Similar to Fanelli \textit{et al.} \cite{fanelli2013}, we also report the mean accuracy as percentage of good estimations (\textit{i.e.}, angle error below \ang{15}).\\ 
The first row of Table \ref{tab:resOurDataset} shows the performance of a baseline system, obtained using the head pose estimation network only, and input depth frames are directly fed to the network without processing and cropping the input around the head. As expected, results are reasonable proving the ability of the deep network to extract useful features for head pose estimation from whole images.\\  
The cropping step is included instead in the other rows, using the ground truth head position as the center and the cropping method described in Section \ref{sec:headlocalization}. 
All three branches (i.e., depth, \textit{FfD}, and Motion Images) of \textit{POSEidon$^+$} framework are individually evaluated. 
In particular, the fifth row includes an indirect evaluation of the reconstruction capabilities of the \textit{Face-from-Depth} network. 
The same network trained and tested on the original gray level images performs similarly to the one trained and tested on \textit{FfD} outputs (Row 3). The similar results confirm that the image reconstruction quality is sufficiently accurate, at least for the pose estimation task.\\  
Results obtained using couples of networks are shown in rows 6 and 7, exploiting concatenation to merge the final layers of each component. 
Finally, the last row reports the performance of the complete framework. To merge layers, we use a \textit{conv} fusion of couples of input types, followed by the \textit{concat} step. We found that it is the best combination, as described in \cite{borghi17cvpr}. Even if the choice of the fusion method has a limited effect (as deeply investigated in \cite{park2016,feichtenhofer2016}), the most significant improvement of the system is reached by combining and exploiting the three input types together.

\begin{figure}[b!]
    \centering
    \includegraphics[width=0.9\linewidth]{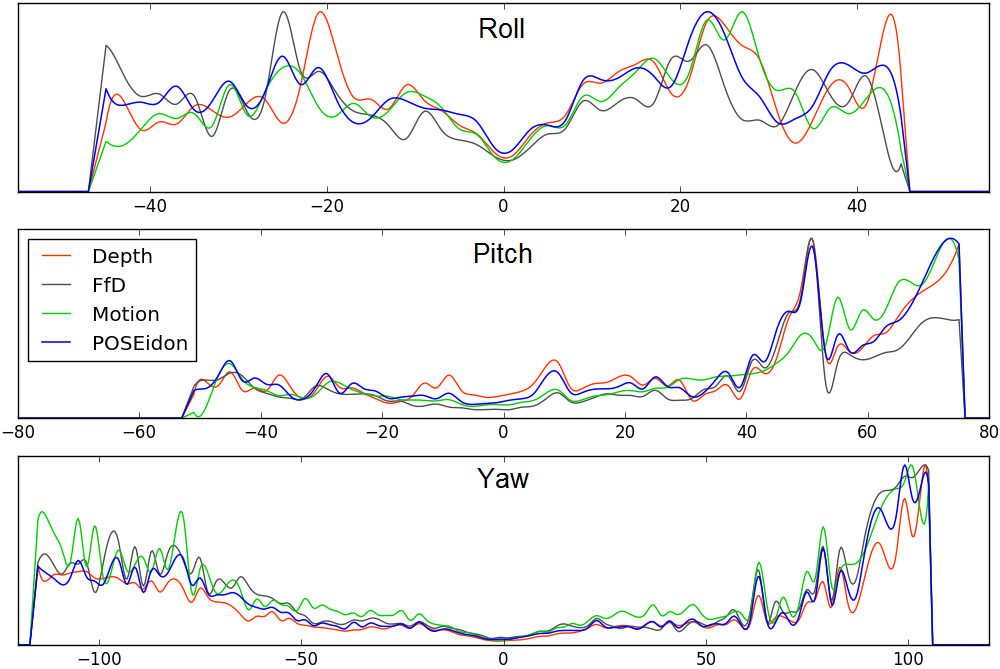}
    \caption{Error distribution of each \textit{POSEidon$^+$} components on \textit{Pandora} dataset. On $x$-axis are reported the ground truth angles, on $y$-axis the distribution of error for each input type.
}
    \label{fig:erroriperangolo}
\end{figure}

Figure \ref{fig:erroriperangolo} shows a comparison of the performance provided by each trident component: each graph plots the error distribution of a specific network with respect to the ground truth value. Depth data allows reaching the lowest error rates for frontal heads, while the other input data types are better in presence of rotated poses. The graphs highlight the averaging capabilities of \textit{POSEidon$^+$} too. \\
Furthermore, in Table \ref{tab:testposeidon} we compare best performance of \textit{POSEidon$^+$} on \textit{Pandora} dataset, obtained exploiting the \textit{FfD} network proposed in this paper and the previous one described in \cite{borghi17cvpr}. We also evaluate \textit{POSEidon$^+$} replacing the central CNN (see Fig. \ref{fig:general}) trained on reconstructed face images with the same CNN but trained on gray-level images (this experiment is here referred as \textit{POSEidon}*). Results confirm that the proposed \textit{POSEidon$^+$} overcomes our preliminary work. The overall quality of reconstructed face images is confirmed and also the feasibility to train and test the pose network on different dataset without a significant drop in performance.

\begin{table}[t!]
\caption{Estimation errors and mean accuracy of the shoulder pose estimation on \textit{Pandora}}
\centering
\small
\begin{tabular}{cc|cccc}
\hline
\multicolumn{2}{c|}{\textbf{Parameters}}  &\multicolumn{3}{c}{\textbf{Shoulders}}   &\multirow{2}{*}{\textbf{Accuracy}}\\
\cline{3-5}
$R_x$  &$R_y$ 		&Pitch		&Roll			&Yaw    &		\\ \hline \hline
\multicolumn{2}{c|}{No crop} &2.5 $\pm$ 2.3			&3.0 $\pm$2.6    &3.7 $\pm$ 3.4 &0.877				\\ \hline
700	 	&250		&2.9 $\pm$ 2.6			&2.6 $\pm$2.5    &4.0 $\pm$ 4.0 	 	&0.845 	\\ \hline
850	 	&250		&2.4 $\pm$ 2.2			&2.5 $\pm$2.2    &3.1 $\pm$ 3.1 	 	&0.911 	\\ \hline
850	 	&500		&\textbf{2.2} $\pm$ \textbf{2.1}			&\textbf{2.3} $\pm$\textbf{2.1}    &\textbf{2.9} $\pm$ \textbf{2.9} 	&\textbf{0.924}	\\ \hline
\end{tabular}
\label{tab:res_crop}
\end{table}

Finally, we compare the results of \textit{POSEidon$^+$} with the state-of-art on the \textit{Biwi} dataset. Due to the lack of a common validation and test protocol, Table \ref{tab:resBiwiPose} is split accordingly to the evaluation procedures adopted, in order to allow fair comparisons. 
For each validation procedure, we report results of \textit{POSEidon$^+$}.
In particular, we implement a 2-folds (half subjects in train and half in test), 4-folds, 5-folds (as adopted in the original works \cite{fanelli2011dagm, fanelli2013}, respectively) and 8-folds subject independent cross evaluations. We also conduct the \textit{Leave-One-Out} (LOO) validation protocol.
We dedicate the last section of Table \ref{tab:resBiwiPose} also for those methods that do not follow a standard evaluation procedure since they create a fixed or random~\cite{ahn2014} sets with a limited number of subjects (or sequences) to test their systems.
Besides, we note that a fair comparison with methods reported in the top part of Table \ref{tab:resBiwiPose} is not possible since they exploit all sequences of \textit{Biwi} dataset for test, while deep learning approaches need a certain amount of training data.\\
Results confirm the excellent performance of \textit{POSEidon$^+$} and the generalization ability across different training and testing subsets with different validation protocol.
The system overcomes all the reported methods, included our previous proposal \cite{borghi17cvpr}. 
The average error is lower than other approaches, even those are not using all the frames available on \textit{Biwi} dataset (some works exclude the frames on which the face detection fails \cite{fanelli2011dagm, fanelli2013}). \\


\noindent \textbf{Shoulder Pose Estimation. }  
The network performing the shoulder pose estimation has been tested on \textit{Pandora} only, due to the lack of the corresponding annotation in the other datasets. Results are reported in Table \ref{tab:res_crop}. \\
In particular, we conduct evaluation on different input types, varying the values $R_x$ and $R_y$ (cf. Section \ref{sec:shoulderpose}) that affect head and shoulder crops. We test also the shoulder pose network using the whole input depth frame, without any crop.
The reported results are very promising, reaching an accuracy of over $92\%$. \\

\begin{table}[t!]
\centering
\small
\caption{Results on \textit{Biwi}, \textit{ICT-3DHP} and \textit{Pandora} dataset of the complete \textit{POSEidon$^+$} pipeline (\textit{i.e.}, head localization, cropping and pose estimation).}
\begin{tabular}{ccccc}
\hline
\multirow{2}{*}{\textbf{Dataset}} 	&\multirow{2}{*}{\textbf{Local.}}  &\multicolumn{3}{c}{\textbf{Head}} \\ \cline{3-5}
 					&	&Pitch			&Roll			&Yaw     			\\ \hline \hline
Biwi 				&3.27$\pm$2.19	& 1.5$\pm$1.4 	&1.6$\pm$1.6    &2.2$\pm$2.0   		\\ \hline
ICT-3DHP 			&-	& 4.9$\pm$4.2 	&3.5$\pm$3.4    &6.8$\pm$6.0  		\\ \hline
Pandora 			&4.27$\pm$3.25	& 7.3$\pm$8.2  	&4.6$\pm$4.5 	&10.3$\pm$11.4 	 	 	\\ \hline
\end{tabular}
\label{tab:pipeline}
\end{table}

\noindent \textbf{Complete pipeline. }  
In order to have a fair comparison, results reported in Tables \ref{tab:resBiwiPose} and \ref{tab:resOurDataset} are obtained using the ground truth head position as input to the crop procedure. 
We finally test the whole pipeline, including the head localization network described in section \ref{sec:headlocalization}, using also \textit{ICT-3DHP} dataset. 
The mean error of the head localization (in pixels) and the pose estimation errors are summarized in Table \ref{tab:pipeline}. Sometimes, the estimated position generates a more effective crop of the head and, as a result, the whole pipeline performs better on the head pose estimation over the \textit{Biwi} dataset. \textit{POSEidon$^+$} reaches valuable results also on the \textit{ICT-3DHP} dataset and it provides comparable results with respect to state-of-the-art methods working on both depth and RGB data (4.9$\pm$5.3, 4.4$\pm$4.6, 5.1$\pm$5.4 \cite{saeed2015}, 7.06, 10.48, 6.90 \cite{baltruvsaitis2012}, for pitch, roll and yaw respectively). We note that \textit{ICT-3DHP} does not include the head center annotation, but the position of the device used to acquire pose data placed on the back of the head, and this partially compromises the performance of our method. Besides, we can not suppose a coherency between the annotations obtained with different IMU devices, in particular regarding the definition of the null position (\textit{i.e.}, when the head angles are equal to zero). 

\noindent The complete framework -- except for the \textit{FfD} module -- has been implemented and tested on a desktop computer equipped with a \textit{NVidia Quadro k2200} GPU board and on a laptop with a \textit{NVidia GTX 860M}. Real-time performance has been obtained in both cases, with a processing rate of more than 30 frames per second. The whole system has been tested instead on a \textit{Nvidia GTX 1080} and is able to run at more than 50 frames per second. 
Some examples of the system output are reported in Figure \ref{fig:demofinale}. 
In addition, the original depth map, the \textit{Face-from-Depth} reconstruction and the motion data given in input to \textit{POSEidon$^+$} are placed on the left of each frame.

\section{Conclusions}
An end-to-end framework to monitor the driver's body pose called \textit{POSEidon$^+$} is presented. In particular, a new \textit{Face-from-Depth} architecture is proposed, based on a Deterministic Conditional GAN approach, to convert depth faces in gray-level images and supporting head pose prediction.\\ 
The system is based only on depth images, no previous computation of specific facial features is required and has shown real-time and impressive results with two public datasets. 
All these aspects make the proposed framework suitable to particular challenging contexts, such as automotive. 
Since the system has been developed with a modular architecture, each module can be used as single or in combination, reaching worst but still satisfactory performances.
This work provides a comprehensive review and a comparison of recent state-of-art works and can be used as a brief review to understanding the current state of the 3D head pose estimation task.
 \ifCLASSOPTIONcompsoc
   \section*{Acknowledgments}
 \else
   \section*{Acknowledgment}
 \fi

This work has been carried out within the projects “Citta educante”  (CTN01-00034-393801)  of  the  National  Technological Cluster on Smart Communities funded by MIUR and “\textit{FAR2015  -  Monitoring  the  car  drivers  attention  with  multisensory  systems,  computer  vision  and  machine  learning}” funded  by  the  University  of  Modena  and  Reggio  Emilia. We also acknowledge \textit{Ferrari} SpA and CINECA for the availability of real car equipments and high performance computing resources, respectively.

\ifCLASSOPTIONcaptionsoff
  \newpage
\fi

\bibliographystyle{IEEEtran}
\bibliography{biblio}

%

\begin{IEEEbiography}
[{\includegraphics[width=1in,height=1.25in,clip,keepaspectratio]{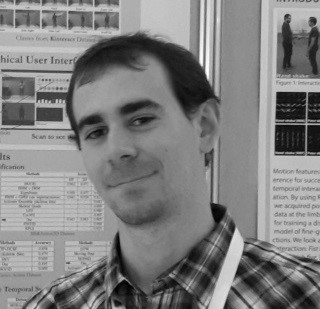}}]{Guido Borghi}
received the master's degree in computer engineering from the University of Modena and Reggio Emilia in 2015. He is currently a Ph.D. candidate within the AImageLab group in Modena. His research topics are about Computer Vision and Deep Learning oriented to Human-Computer Interaction applications with 3D data. 
\end{IEEEbiography}

\begin{IEEEbiography}
[{\includegraphics[width=1in,height=1.25in,clip,keepaspectratio]{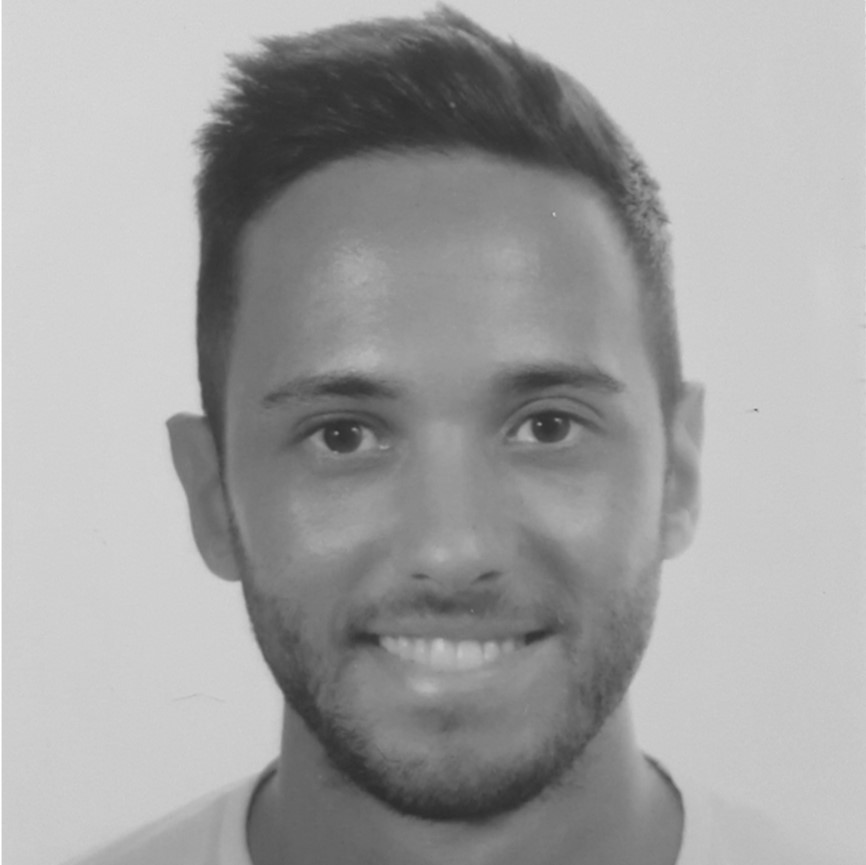}}]{Matteo Fabbri}
is currently a Ph.D. student at the International Doctorate School in ICT of the University of Modena e Reggio Emilia. He works under the supervision of Prof. Rita Cucchiara and Ing. Simone Calderara, on computer vision and deep learning for people behavior understanding. His research interests include generative models and multiple object tracking.
\end{IEEEbiography}

\begin{IEEEbiography}
[{\includegraphics[width=1in,height=1.25in,clip,keepaspectratio]{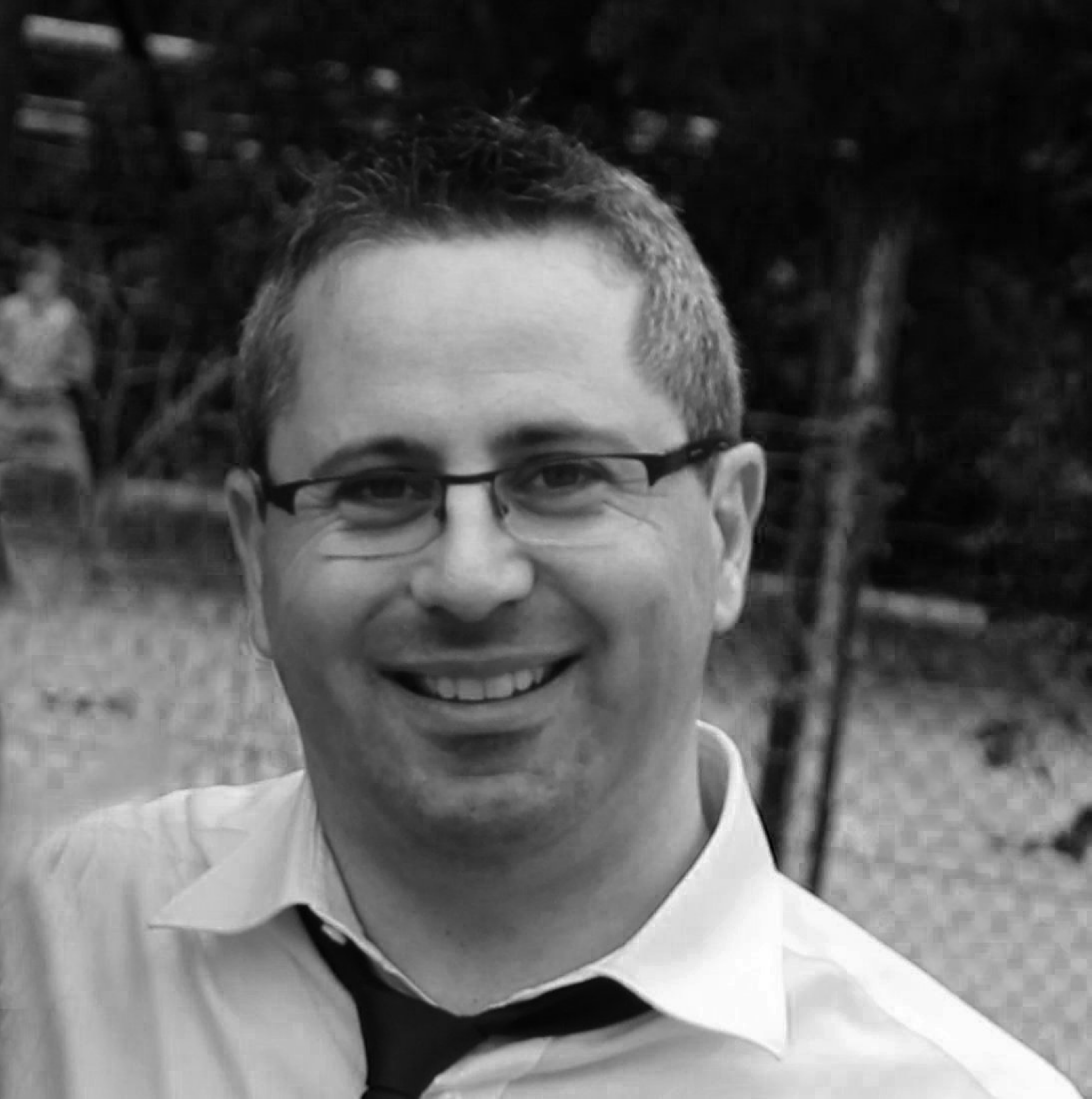}}]{Roberto Vezzani}
Roberto Vezzani graduated in Computer Engineering in 2002 and received his Ph.D. course in Information Engineering in 2007 at the University of Modena and Reggio Emilia, Italy.
Since 2016 is Associate Professor.
His research interests mainly belong to computer vision systems for human computer interaction, video surveillance, with a particular focus on motion detection, people tracking and re-identification.  He is a member of ACM, IEEE, and IAPR.
\end{IEEEbiography}

\begin{IEEEbiography}
[{\includegraphics[width=1in,height=1.25in,clip,keepaspectratio]{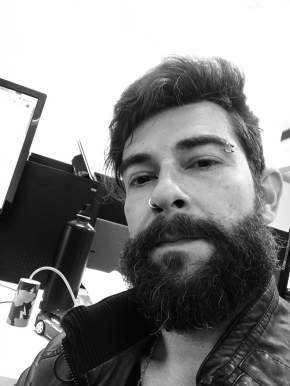}}]{Simome Calderara}
received a computer engineering masters degree in 2005 and the Ph.D. degree in 2009 from the University of Modena and Reggio Emilia, where he is currently an assistant professor within the AImageLab group. His current research interests include computer vision and machine learning applied to human behavior analysis, visual tracking in crowded scenarios, and time series analysis for forensic applications. He is a member of the IEEE.
\end{IEEEbiography}

\begin{IEEEbiography}
[{\includegraphics[width=1in,height=1.25in,clip,keepaspectratio]{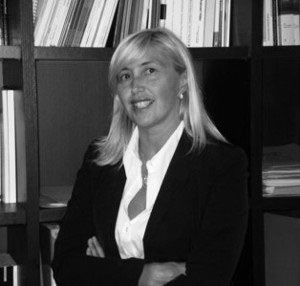}}]{Rita Cucchiara}
received the master’s degree in Electronic Engineering and the Ph.D. degree in Computer Engineering from the University of Bologna, Italy, in 1989 and 1992, respectively. Since 2005, she is a full professor at the University of Modena and Reggio Emilia, Italy, where she heads the AImageLab group and is Director of the SOFTECH-ICT research center. She is currently President of the Italian Association of Pattern Recognition, (GIRPR), affiliated with IAPR. She published more than 300 papers on pattern recognition computer vision and multimedia, and in particular in human analysis, HBU, and egocentric-vision. The research carried out spans on different application fields, such as video surveillance, automotive and multimedia big data annotation. Currently, she is AE of IEEE Transactions on Multimedia and serves in the Governing Board of IAPR and in the Advisory Board of the CVF.
\end{IEEEbiography}

\end{document}